\documentclass{bmvc2k}

\usepackage{epsfig}
\usepackage{graphicx}
\usepackage{amsmath}
\usepackage{amssymb}
\usepackage{ctable}
\usepackage{capt-of}

\title{Visibility-aware Multi-view Stereo Network}

\addauthor{Jingyang Zhang}{jzhangbs@cse.ust.hk}{1}
\addauthor{Yao Yao}{yyaoag@cse.ust.hk}{1}
\addauthor{Shiwei Li}{sli@altizure.com}{2}
\addauthor{Zixin Luo}{zluoag@cse.ust.hk}{1}
\addauthor{Tian Fang}{fangtian@altizure.com}{2}

\addinstitution{
 The Hong Kong University of Science and Technology\\
 Hong Kong SAR, China
}
\addinstitution{
 Everest Innovation Technology\\
 Hong Kong SAR, China
}

\runninghead{Zhang, Yao, Li, Luo, Fang}{Vis-MVSNet}

\begin{document}

\maketitle

\begin{abstract}
	Learning-based multi-view stereo (MVS) methods have demonstrated promising results. However, very few existing networks explicitly take the pixel-wise visibility into consideration, resulting in erroneous cost aggregation from occluded pixels. In this paper, we explicitly infer and integrate the pixel-wise occlusion information in the MVS network via the matching uncertainty estimation. The pair-wise uncertainty map is jointly inferred with the pair-wise depth map, which is further used as weighting guidance during the multi-view cost volume fusion. As such, the adverse influence of occluded pixels is suppressed in the cost fusion. The proposed framework \textit{Vis-MVSNet} significantly improves depth accuracies in reconstruction scenes with severe occlusion. Extensive experiments are performed on \textit{DTU}, \textit{BlendedMVS}, and \textit{Tanks and Temples} datasets to justify the effectiveness of the proposed framework. 
\end{abstract}

\section{Introduction}
Multi-view Stereo (MVS) is one of the core problems in computer vision, which is essential to a variety of applications including image-based 3D modeling, city-scale survey and autonomous driving. While the problem is mainly solved by classical methods \cite{campbell2008using, furukawa2009accurate, tola2012efficient, galliani2015massively, schonberger2016pixelwise}, recent learning-based methods \cite{yao2018mvsnet,yao2019recurrent,gu2020cascade} have also shown competitive results compared to previous state-of-the-arts. Learning-based methods usually extract deep image features from input images, which implicitly introduces global semantic such as specularity and reflection priors during the reconstruction process. Moreover, MVS networks usually apply 3D convolution neural networks (CNNs) for the cost volume regularization, which is more powerful than engineered cost regularization in classical methods.

One critical factor in MVS is the pixel-wise visibility: whether a 3D point is visible in given images. However, such visibility information is unknown before the 3D model is densely recovered, which implies a chicken-and-egg problem. In traditional MVS algorithms, the visibility issue is well understood: some approaches simply reject patch pairs according to pre-determined criteria, and then update the cost aggregation with only the inlier patch pairs \cite{furukawa2009accurate,tola2012efficient,xu2019multi}. More advanced approaches, such as COLMAP \cite{zheng2014patchmatch,schonberger2016pixelwise}, compute the visibility information and aggregate the pair-wise matching cost based on a probabilistic framework, where visibility and depth are alternatively updated in E-step and M-step. 

However, for current learning-based MVS methods, very few of them have acknowledged this problem and have explicitly handled the visibility issue. For example, MVSNet and its following works \cite{yao2018mvsnet,yao2019recurrent,chen2019point,gu2020cascade,cheng2020deep,yang2020cost} feed multi-view features from all views into a variance-based cost metric regardless of the visibility of the pixel. Other methods apply either averaging \cite{hartmann2017learned} or max pooling \cite{huang2018deepmvs} to aggregate the matching cost. While it is possible that the network could implicitly learn how to discard the invisible views for each pixel, the unsolved visibility problem may inevitably deteriorate the final reconstruction.

In this work, we present an end-to-end network architecture that takes pixel-wise visibility information into account. The depth map is estimated from multi-view images in a two-step manner.
First, matching is performed for each reference-source image pair and a latent volume representing the pair-wise matching quality is obtained. This volume further regresses to an intermediate estimation of a depth map and an uncertainty map, where the uncertainty is transformed from the depth-wise entropy of the probability volume. 
Second, to attenuate unmatchable pixels, we fuse all pair-wise latent volumes to one multi-view cost volume by using pair-wise matching uncertainties as weighting guidance. The fused volume is regularized and regresses to the final depth estimation. 
We also integrate several practical components from recent MVS networks, including group-wise correlation and \cite{guo2019group} coarse-to-fine strategy \cite{gu2020cascade} to further boost the overall reconstruction quality. 
Our network is end-to-end trainable and the uncertainty part is trained in an unsupervised manner. In this case, we can directly utilize existing MVS datasets with only ground truth depth maps to train the visibility-aware MVS network.

The proposed Vis-MVSNet is evaluated on \textit{DTU} \cite{jensen2014large} and \textit{BlendedMVS} \cite{yao2020blendedmvs} datasets and is benchmarked on \textit{Tanks and Temples} \cite{knapitsch2017tanks} dataset. Our method ranks $1^{st}$ among all submissions in the \textit{Tanks and Temples} online benchmark (until May 1, 2020). Comparisons with previous methods and ablation studies in the experiment section demonstrate the significant improvement bought by our approach, especially when the occlusion problem is severe in input images.

\section{Related Work}

\begin{figure}[]
	\centering
	\includegraphics[width=\textwidth]{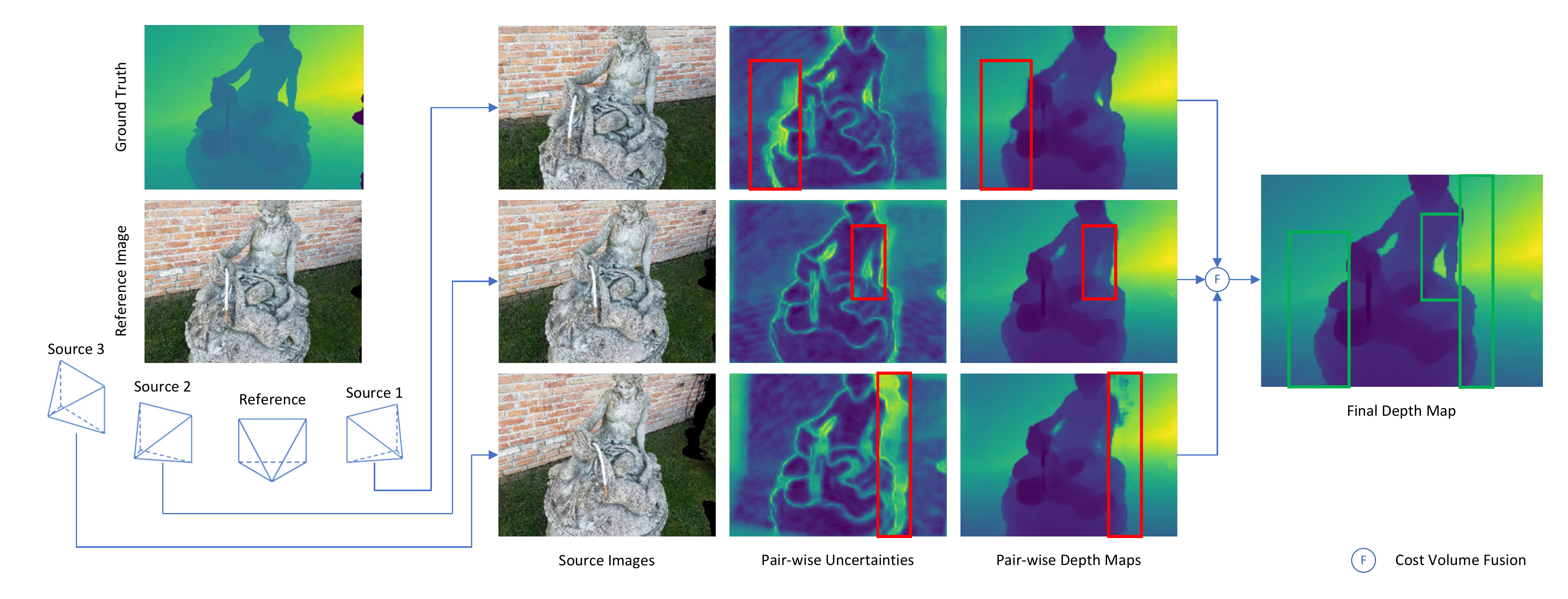}
	\vspace{-8mm}
	\caption{Illustration of the visibility-awared fusion. For each reference-source pair, the uncertainty map successfully estimates the visibility of the pixels, and the depths of the occluded pixels are not correct. During the fusion, the occluded pixels are attenuated, resulting in a well reconstructed final depth map. }
	\vspace{-5mm}
	\label{fig:method2}
\end{figure}

\paragraph{Learning-based MVS}
Learning-based methods have shown great potentials to replace each step in traditional MVS reconstructions. The learnable multi-view cost metric \cite{hartmann2017learned} is first proposed to measure the multi-view photo-consistency between image patches. Later, SurfaceNet \cite{ji2017surfacenet} is proposed to learn the cost volume regularization from geometry ground truth. The authors of LSM \cite{kar2017learning} apply the differentiable projection in the network and propose the first end-to-end learnable network for low-resolution MVS reconstruction. DeepMVS \cite{huang2018deepmvs} reprojects images to 3D plane-sweeping volumes, performs intra-volume aggregation, and applies inter-volume aggregation to fuse the volumes and generate the depth map output. RayNet \cite{paschalidou2018raynet} encodes the camera projection to the network, and utilizes the Markov Random Field to predict the surface label. 

Another recent popular network for MVS reconstruction is MVSNet \cite{yao2018mvsnet}. MVSNet first extracts deep image features and then warps these features into the reference camera frustum to build a cost volume via differentiable homographies. To reduce the memory consumption during the network inference, the follow-up R-MVSNet \cite{yao2019recurrent} replaces the 3D CNNs regularization module with a 2D GRU recurrent network. 
Point-MVSNet \cite{chen2019point} proposes a point-based depth map refinement network to improve the output accuracy and MVS-CRF \cite{xue2019mvscrf} introduces the conditional
random field optimization during the depth map estimation. More recently, CasMVSNet \cite{gu2020cascade}, CVP-MVSNet \cite{yang2020cost} and UCSNet \cite{cheng2020deep} integrate the coarse-to-fine strategy to the learning-based MVS reconstruction. These works preserve an image feature pyramid and generate an initial depth estimation with large depth interval at a low resolution. In following stages, cost volumes are constructed with a narrow depth range centering at the depth estimation from previous stages. The coarse-to-fine architecture successfully reduces memory consumption so that they support deeper backbone networks and higher resolution outputs. However, these methods all apply a variance-based cost metric, which is under the assumption that a given pixel is visible in all input images. As a result, an increasing number of input images would lead to even a worse depth map estimation quality.

\vspace{-5mm}\paragraph{Visibility Estimation}
Visibility estimation is a well-acknowledged problem in classic MVS reconstructions. Previous works include heuristic cost thresholding methods \cite{furukawa2009accurate,tola2012efficient,xu2019multi} and more complicated joint depth-visibility estimation methods \cite{zheng2014patchmatch, schonberger2016pixelwise}. For latter approaches, the per-pixel visibility is usually jointly recovered during the depth map estimation process through an EM-based method. However, these methods apply a probabilistic framework which is hard to be directly integrated with deep neural networks. To handle the visibility issue in the learning-based frameworks, we should consider other alternatives for joint depth map and visibility estimation.

Current deep learning methods take visibility into account in an implicit manner. MVSNet \cite{yao2018mvsnet} reduces the feature volumes from different source views by variance metric which considers each view equally and claims that information from invisible pixels can be filtered out in the regularization. Such implicit method heavily relies on the regularization of the neural network. Besides, DeepMVS \cite{huang2018deepmvs} applies max pooling of multiple feature volumes to select the best latent representation, which is expected to be generated from a matchable pair. However, the fused volume is only related to the information from the best view, which loses the advantage of MVS that a more robust prediction can be produced by multiple observation. Instead, we start from pair-wise cost volumes to identify the pair-wise matching quality, and fuse the pair-wise volumes by weighted sum where weights of unmatchable pairs are reduced. 

\vspace{-5mm}\paragraph{Uncertainty Estimation}
In our approach, visibility is indicated by the matching uncertainty of the pair-wise depth map. Uncertainty (or confidence) estimation for two-view depth or disparity estimation has been widely studied for classic methods by Hu and Mordohai \cite{hu2012quantitative}. The majority of such methods examine the properties of the probability distribution over all the depth or disparity hypotheses. End-to-end deep neural networks \cite{poggi2016learning, kim2018unified, tosi2018beyond, kim2019laf} are also applied to estimate the uncertainty map for two-view stereo. Recently, Kendall and Gal \cite{kendall2017uncertainties} propose to jointly estimate the network output and its uncertainty based on the Bayesian neural network. However, this method cannot be directly adopted in our framework because they operate on 2D outputs, while we believe that it is more reasonable to estimate uncertainty from the 3D probability volume. Therefore we follow \cite{zhang2020learning} to use the depth-wise entropy of the probability volume to explicitly measure the pair-wise matching uncertainty. 


\begin{figure}[]
	\centering
	\includegraphics[width=\textwidth]{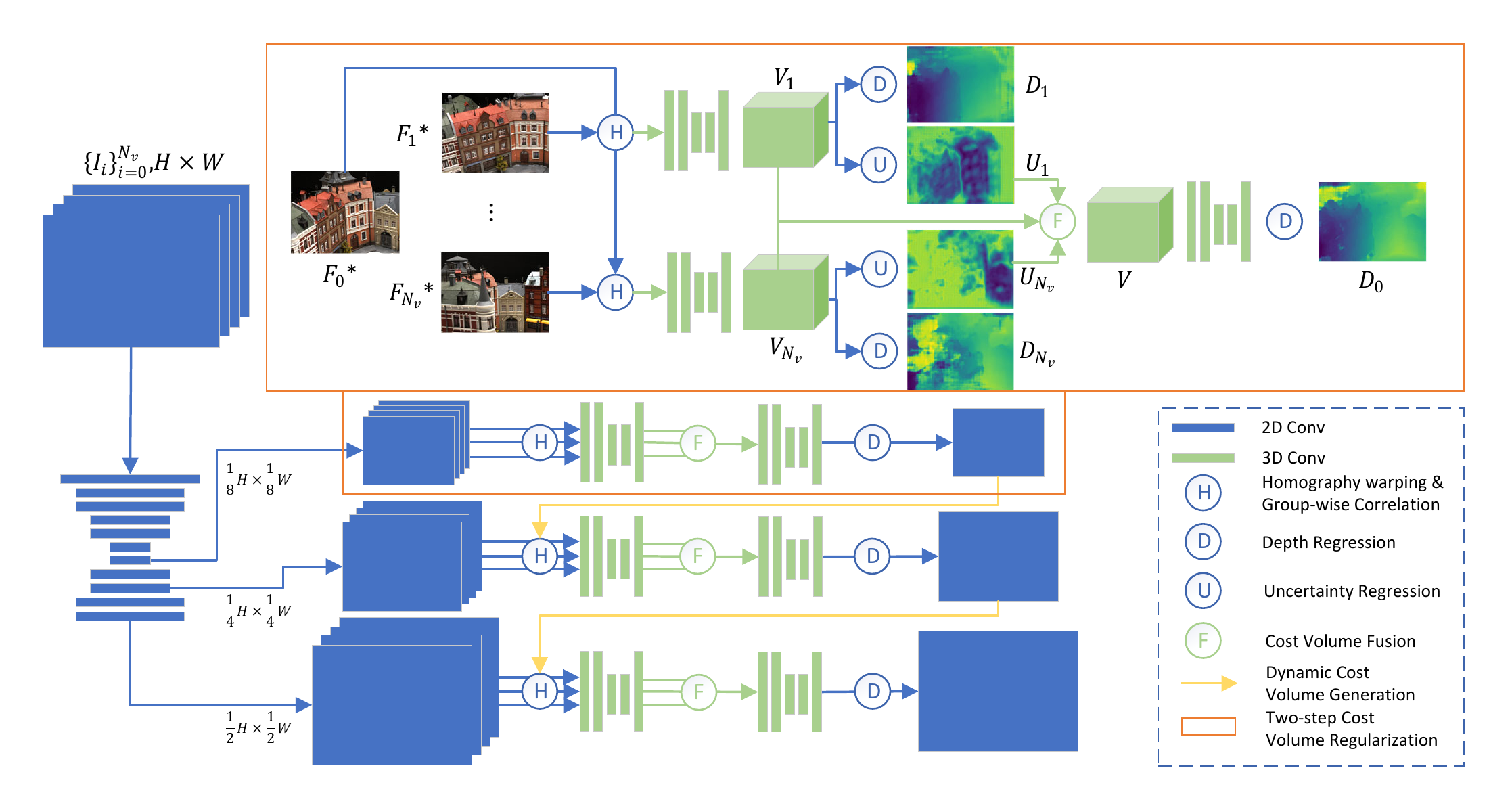}
	\vspace{-8mm}
	\caption{The proposed framework. For every reference-source pair, we jointly infer the depth map and the uncertainty map. The latent volumes are fused according to the uncertainty. And the fused volume is further regularized for the final depth map regression. *The feature maps. The images here only show the original image of the feature maps. }
	\vspace{-3mm}
	\label{fig:framework}
\end{figure}

\section{Method}

\subsection{Overview}\label{method_overview}
The outline of the framework is illustrated in Fig.\ \ref{fig:framework}. Given a reference image $\mathbf{I}_0$ and a set of neighboring source images $\{\mathbf{I}_i\}^{N_v}_{i=1}$, the framework predicts a reference depth map $\mathbf{D}_{0}$ aligned with $\mathbf{I}_0$. In our network, we apply the coarse-to-fine depth estimation strategy as recent networks \cite{gu2020cascade}. First, all images are fed into a 2D UNet \cite{ronneberger2015u} which extracts the multi-scale image features. The extracted features at the last three scales in the decoder part are preserved and will be used to construct cost volumes at three different resolutions. For the reconstruction at the $k$-th stage, the cost volume will be regularized and produce a depth map $\mathbf{D}_{k,0}$ with the same resolution to the input feature map. Intermediate depth maps from previous stages will be used for the cost volume construction at next stages and $\mathbf{D}_{3,0}$ will be served as the final output $\mathbf{D}_{0}$ of the system.

The network details within the $k$-th stage are described as follows. First, pair-wise cost volumes are constructed for each reference-source pairs. 
For the $i$-th pair, by assuming that the reference image has depth $d$, we can obtain a reprojected feature map ${\mathbf{F}}_{k,i\rightarrow 0}(d)$ from the source view. The groupwise correlation \cite{guo2019group} between the reference and the warped source feature map is calculated as the cost map. Then the cost maps for all the depth hypothesis are stacked together as the cost volume. The resulting cost volume $\mathbf{C}_{k,i}$ of the $i$-th image pair in the $k$-th stage is of size $N_{d,k} \times H \times W \times N_c$, where $N_{d,k}$ is the depth hypothesis number in the $k$-th stage and $N_c = 8$ is the group number of the group-wise correlation operation.
The set of the hypotheses is predetermined for the first stage, and is dynamically determined for the second and third stages according to the depth map output of the previous stage. The calculation of the dynamic depth range will be explained in Sec.\ \ref{method_cas}. 

The regularization of the cost volume consists of two steps. First, every pair-wise cost volume is regularized to a latent volume $\mathbf{V}_{k,i}$ separately. Then, all latent volumes are fused to $\mathbf{V}_k$ which is further regularized to probability volume $ \mathbf{P}_k $ and regresses to the final depth map of the current stage $\mathbf{D}_{k,0}$ via \textit{soft-argmax} \cite{kendall2017end} operation. The fusion of the latent volumes is visibility-awared. First, we measure the visibility by jointly inferring pair-wise depth and uncertainty. Each latent volume is transformed to a probability volume $\mathbf{P}_{k,i}$ through additional 3D CNNs and the \textit{softmax} operation. The depth map $\mathbf{D}_{k,i}$ and the uncertainty map $ \mathbf{U}_{k,i} $ are jointly inferred via \textit{soft-argmax} and \textit{entropy} operation, which will be explained in Sec.\ \ref{method_uncertainty}. Then the uncertainty maps join the volume fusion as the weighting guidance, which is further described in Sec.\ \ref{method_fusion}.

\subsection{Uncertainty Estimation}\label{method_uncertainty}
In current learning-based MVS, the depth map is usually regressed from probability volume via the \textit{soft-argmax} operation. For simplicity, the stage number $ k $ is omitted below. We denote the probability distribution over all the depth hypotheses as $ \{ \mathbf{P}_{i, j} \}_{j=1}^{N_d} $. The \textit{soft-argmax} operation is equivalent to computing the expectation of this distribution and $ \mathbf{D}_i $ is computed as:

\begin{equation}
\mathbf{D}_{i} = \sum_{j=1}^{N_d} d_j \mathbf{P}_{i, j}
\end{equation}

To jointly regress the depth estimation and its uncertainty, we assume that the depth estimation follows the Laplacian distribution \cite{kendall2017uncertainties}. In this case, the estimated depth and the uncertainty maximize the likelihood of the observed ground truth: $p( \mathbf{D}_{gt, i} | \mathbf{D}_{i}, \mathbf{U}_{i} ) = 1/(2\mathbf{U}_{i}) \cdot \exp( |\mathbf{D}_{i} - \mathbf{D}_{gt, i}| / \mathbf{U}_{i} )$. Notice that the probability distribution $\{ \mathbf{P}_{i, j} \}_{j=1}^{N_d}$ also reflects the matching quality. We thus apply the entropy map $\mathbf{H}_{i}$ of $\{ \mathbf{P}_{i, j} \}_{j=1}^{N_d}$ to measure the depth estimation quality. And the uncertainty map $ \mathbf{U}_{i} $ is transformed from $\mathbf{H}_{i}$ by a function $f_u$, which is presented as a shallow 2D CNN in the network. 
\begin{equation}
\mathbf{U}_{i} = f_u (\mathbf{H}_{i}) = f_u(\sum_{j=1}^{N_d} - \mathbf{P}_{i, j} \log \mathbf{P}_{i, j})
\end{equation}
The reason of adopting the entropy is that the randomness of the distribution is negatively related to the uni-modal distribution. And the uni-modality is an indicator of high confidence.

To jointly learn the depth map estimation $\mathbf{D}_{i}$ and its uncertainty $\mathbf{U}_{i}$, we minimize the negative log likelihood described above. 
\begin{equation}\label{eq:joint_loss}
\begin{split}
L_{i}^{joint} &= \frac{1}{|I_0^{valid}|} \sum_{x\in I_0^{valid}} -\log (\frac{1}{2\mathbf{U}_{i}}\exp \frac{| \mathbf{D}_{i} - \mathbf{D}_{gt, i} |}{\mathbf{U}_{i}}) \\
&=\frac{1}{|I_0^{valid}|} \sum_{x\in I_0^{valid}} \frac{1}{\mathbf{U}_{i}}  | \mathbf{D}_{i} - \mathbf{D}_{gt, i} | + \log \mathbf{U}_{i}
\end{split}
\end{equation}
Constants are omitted in the formula. For numerical stability, in practice we infer $ \mathbf{S}_{i} = \log \mathbf{U}_{i} $ instead of $ \mathbf{U}_{i} $ directly. The log uncertainty map $ \mathbf{S}_{i} $ is also transformed from the entropy map $ \mathbf{H}_{i} $ by a shallow 2D CNN. 

The loss can also be interpreted as applying attenuation to the $ L_1 $ loss between the estimation and the ground truth with a regularization term. The intuition is that the interference from the erroneous samples should be reduced. 

\subsection{Volume Fusion}\label{method_fusion}
In this section we introduce the visibility-aware volume fusion. For simplicity, the stage number $ k $ is omitted. Given the pair-wise latent cost volumes $ \{ \mathbf{V}_i \}_{i=1}^{N_v} $, a single volume $ \mathbf{V} $ is fused from the volumes by weighted sum, where the weight is negatively related to the estimated pair-wise uncertainty. 
\begin{equation}\label{eq_method_general2}
\mathbf{V} = (\sum_{i=1}^{N_v} \frac{1}{\exp \mathbf{S}_i} )^{-1} \sum_{i=1}^{N_v} (\frac{1}{\exp \mathbf{S}_i} \mathbf{V}_i)
\end{equation}
The pixels with large uncertainty are more likely to be located in the occluded regions, and thus the values in the latent volume should be attenuated. The attenuation scale is chosen to be identical with the one in the joint loss (Eq.\ \ref{eq:joint_loss}). 

An alternative to the weighted sum is applying threshold for $ \mathbf{S}_i $ and perform a hard visibility selection for each pixel. However, lacking an interpretation of the value $ \mathbf{S}_i $, we can only have an empirical threshold that may not be universal. Instead, the volumes are summed with normalized weight, which considers $ \mathbf{S}_i $ in a relative manner. 

\subsection{Coarse-to-fine Architecture}\label{method_cas}
Our coarse-to-fine architecture mainly follows the recent Cas-MVSNet \cite{gu2020cascade}. In all the stages, depth hypothesis are uniformly sampled from a depth range. The first stage takes image features at low resolution and constructs cost volume with the predetermined depth range but larger depth interval, while the following stages use high spatial resolution, narrower depth range and smaller depth interval. 

For the first stage, the depth range is $ [d_{min}, d_{min}+2\Delta d) $ and the depth number is $ N_{d,1} $, where $ d_{min} $, $ \Delta d $ and $ N_{d,1} $ is predetermined. For the $ k $-th stage ($ k \in \{2,3\} $), the depth range, sample number and interval are reduced. And the ranges are centered at the depth estimation from the previous stage, which are different for each pixels. The depth range for pixel $ x $ is $ [\mathbf{D}_{k-1,0}-w_k\Delta d, \mathbf{D}_{k-1,0}+w_k\Delta d) $ and the depth number is $ p_kN_{d,k} $, where $ w_k<1 $ and $ p_k<1 $ are the predefined scaling factors, and $ \mathbf{D}_{k-1,0} $ is the final depth estimation of pixel $ x $ from the last stage $ k-1 $. 

\subsection{Training Loss}\label{method_loss}
For each stage, the loss is the combination of the pair-wise $ L_1 $ loss, the pair-wise joint loss and the $ L_1 $ loss of the final depth map. And the total loss is the weighted sum of the loss from three stages. For all the losses derived from the absolute difference between the estimation and the ground truth, the per-pixel differences are divided by the depth interval of the final stage. 
\begin{equation}
L = \sum_{k=1}^3\lambda_k[L_{1,k}^{final} + \frac{1}{N_v} \sum_{i=1}^{N_v} (L_{1, k,i}^{pair} + L_{k,i}^{joint})]
\end{equation}
The pair-wise $ L_1 $ losses are included because the uncertainty loss tends to over-relax the pair-wise depth and uncertainty estimation. The pair-wise $ L_1 $ losses here could guarantee a qualified pair-wise depth map estimation.


\begin{figure}[]
	\centering
	\resizebox{\textwidth}{!}{%
		\begin{tabular}{@{\hskip2pt}c@{\hskip2pt}@{\hskip2pt}c@{\hskip2pt}@{\hskip2pt}c@{\hskip2pt}@{\hskip2pt}c@{\hskip2pt}}
			\includegraphics[width=.15\linewidth]{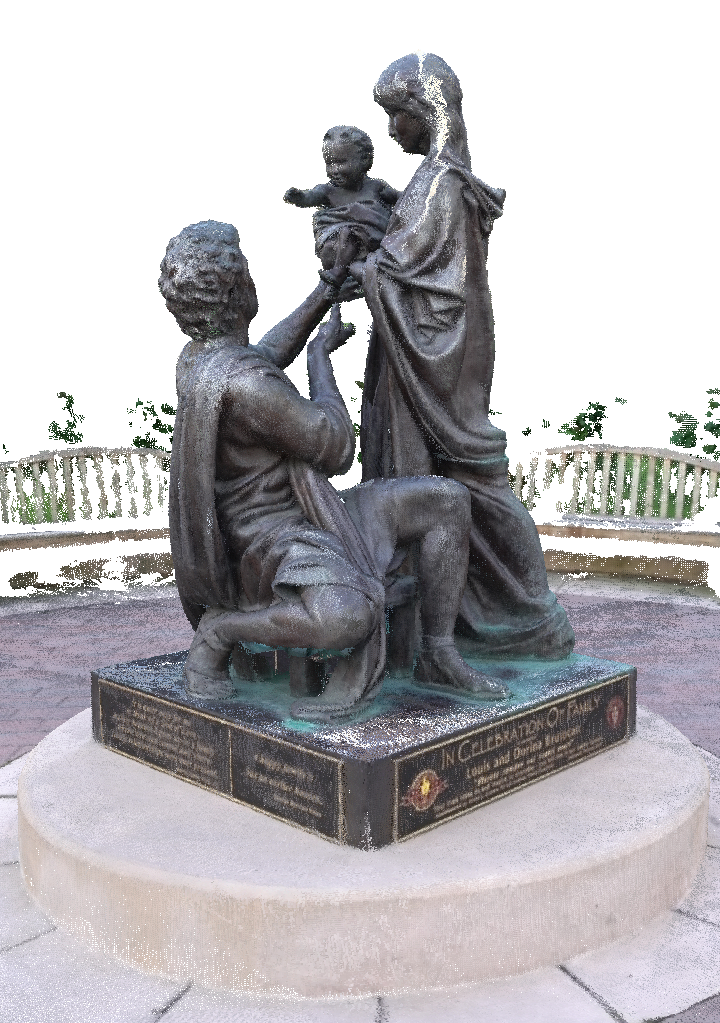} &
			\includegraphics[width=.30\linewidth]{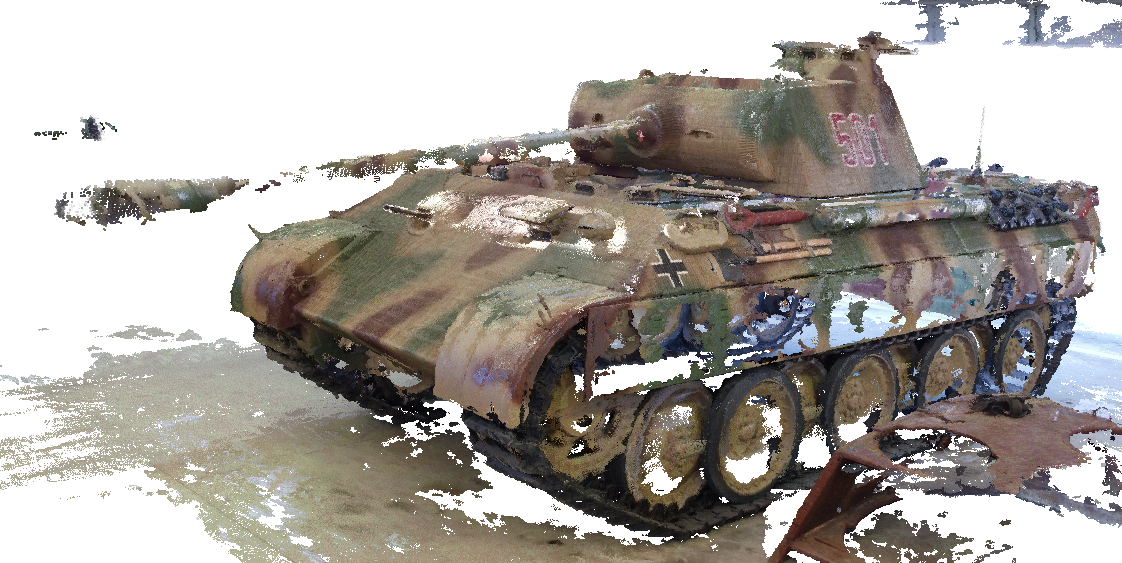} &
			\includegraphics[width=.20\linewidth]{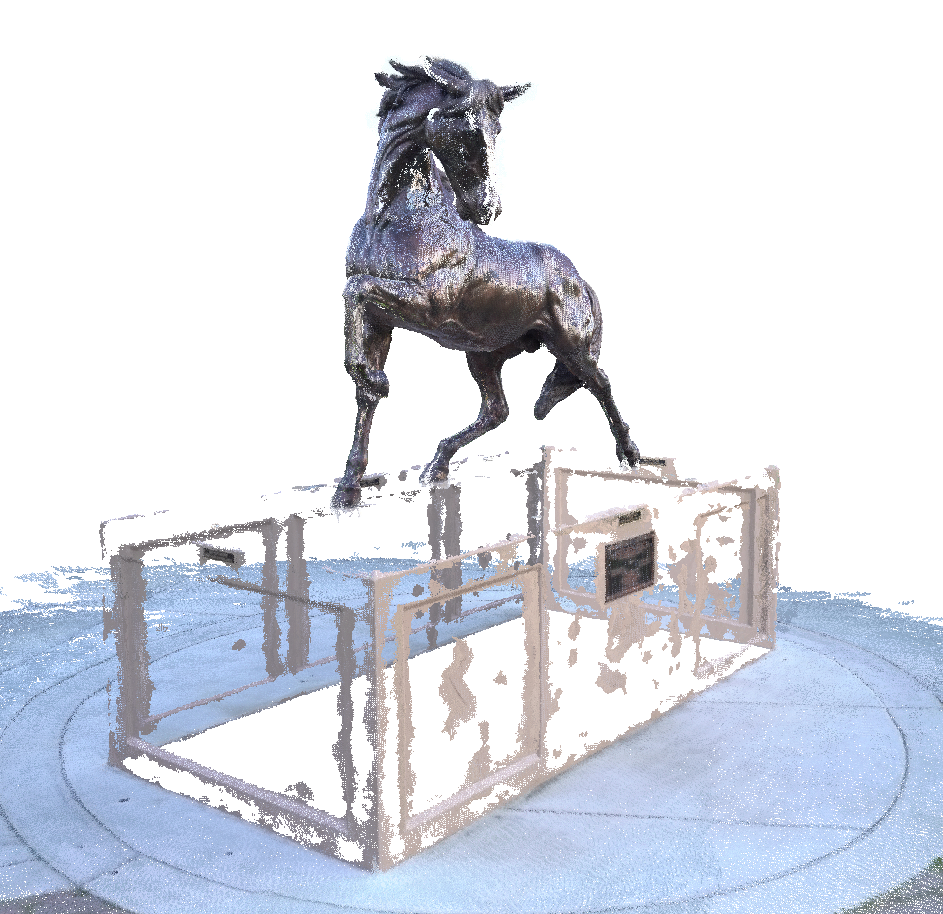} &
			\includegraphics[width=.25\linewidth]{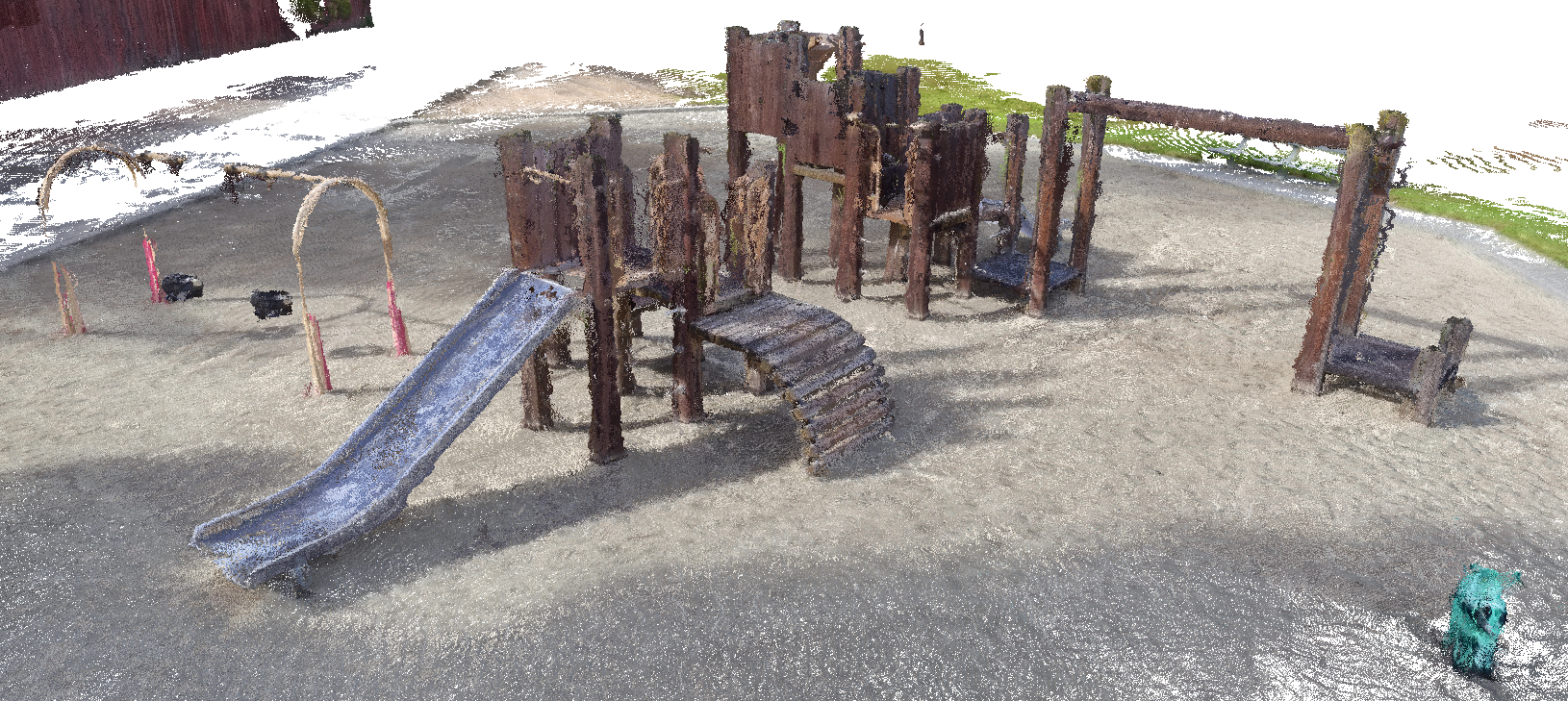} \\
			Family & Panther & Horse & Playground \\
			\includegraphics[width=.1\linewidth]{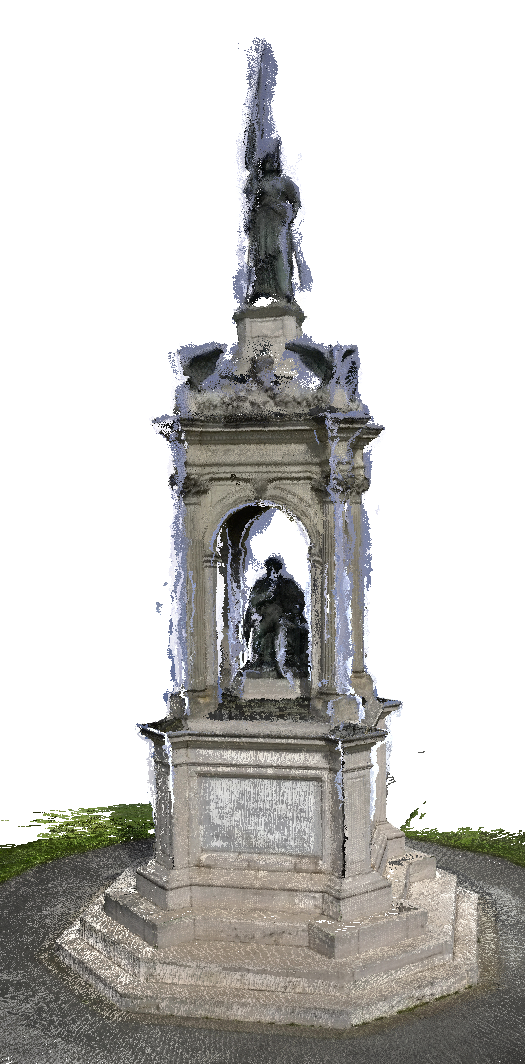} &
			\includegraphics[width=.27\linewidth]{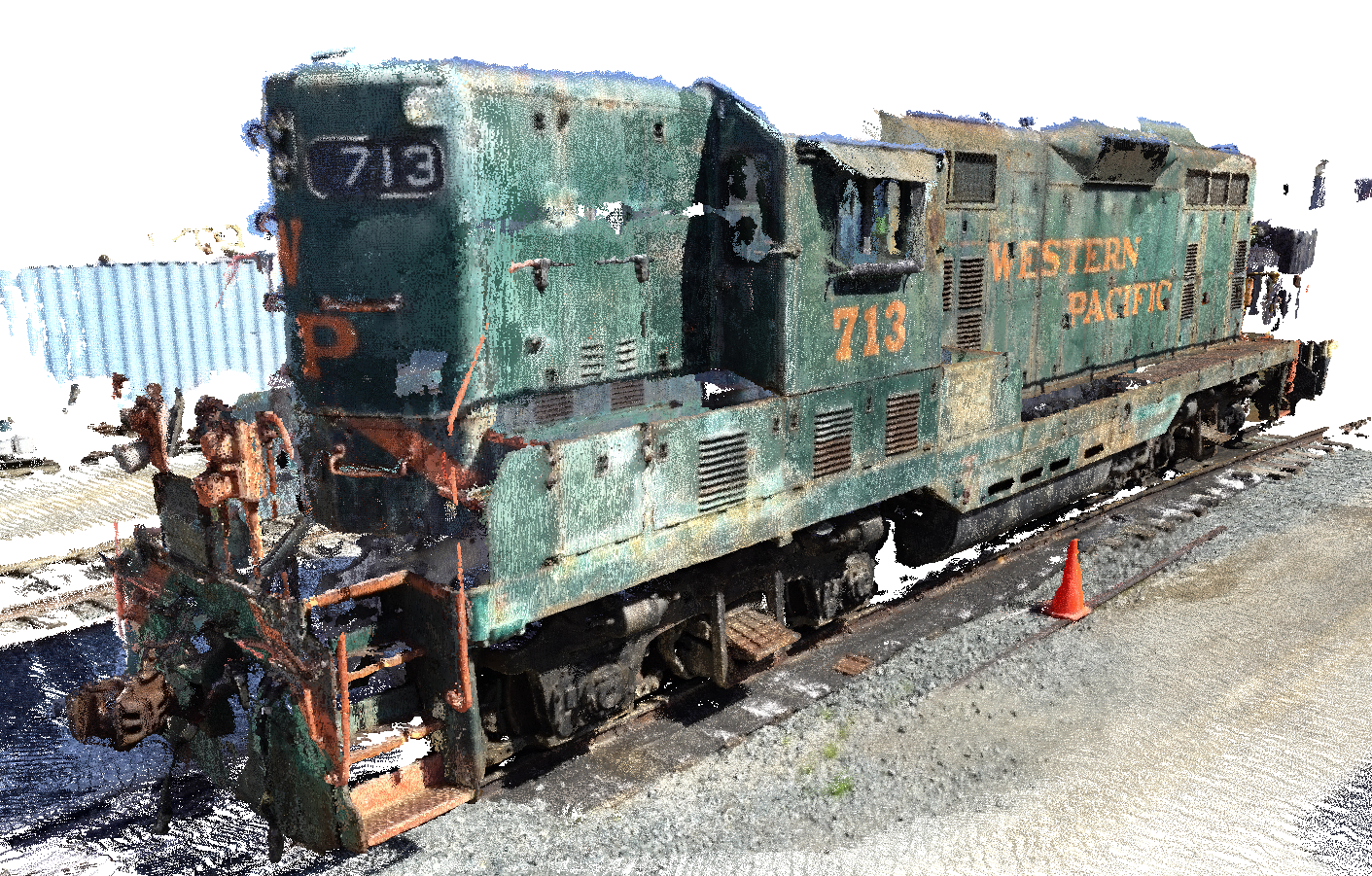} &
			\includegraphics[width=.18\linewidth]{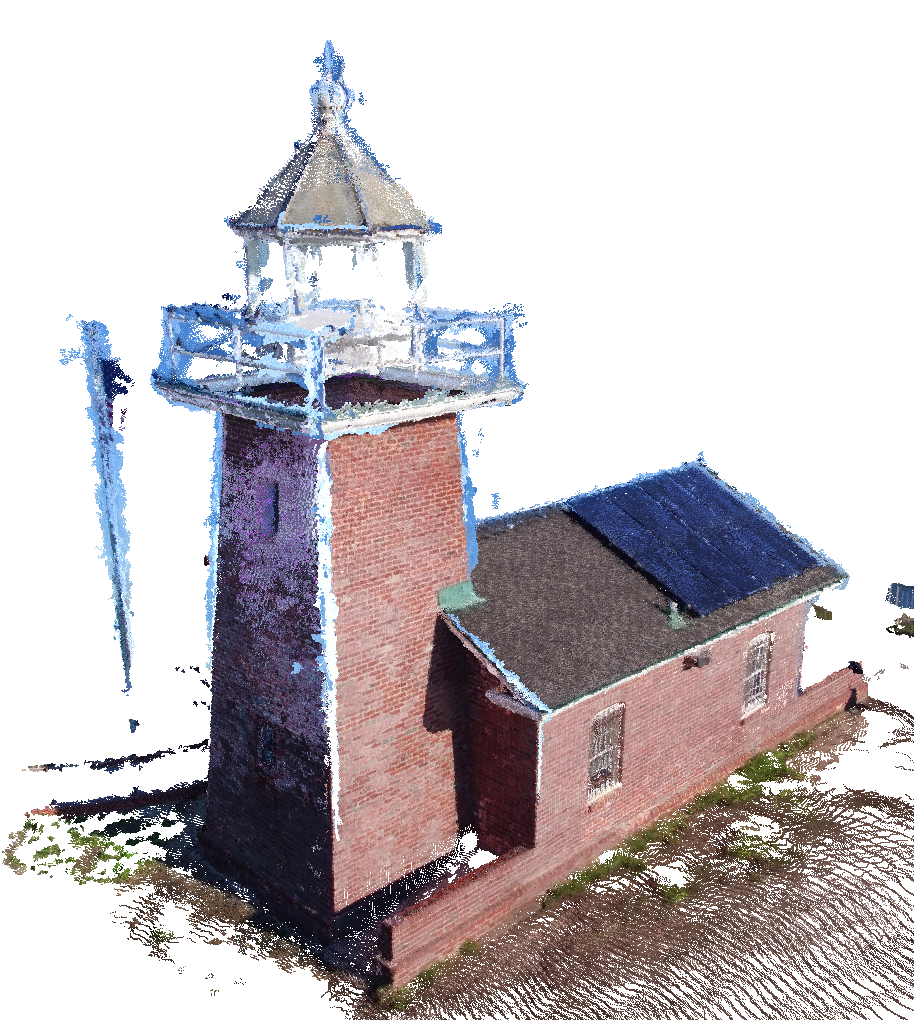} &
			\includegraphics[width=.27\linewidth]{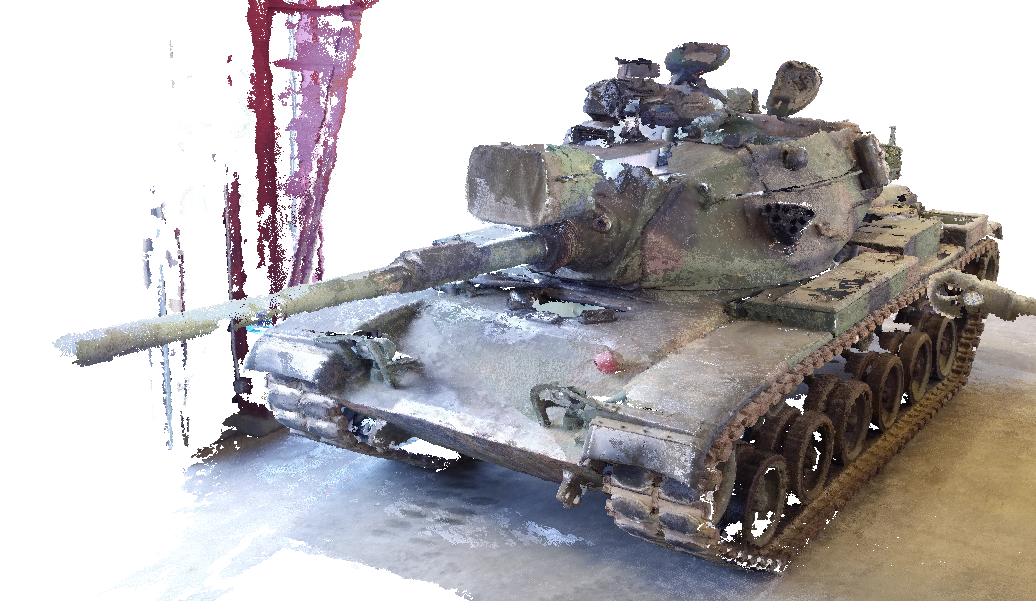} \\
			Francis & Train & Lighthouse & M60
		\end{tabular}
	}
	\caption{Qualitative result of the point cloud on the \textit{intermediate set} of \textit{Tanks and Temples}. }
	\vspace{-3mm}
	\label{fig:tanks}
\end{figure}

\section{Experiment}

\subsection{Implementation}\label{sec:implementation}
\paragraph{Training}
Our network is trained on \textit{BlendedMVS} \cite{yao2020blendedmvs} training set for most experiments (Sec. \ref{sec:tnt} and \ref{sec:ablation}) and is trained on DTU training set \cite{jensen2014large} for DTU benchmarking (Sec. \ref{sec:dtu}). For both training sets, we use the input image size of $640\times512$ and output depth map size of $320\times256$. Source images for the given reference are selected as previous methods \cite{yao2018mvsnet,yao2019recurrent} and we set the number of source views to $N_v=3$ during training. For depth samples at different stages, we set the depth hypothesis numbers to $ N_{d,1}, N_{d,2}, N_{d,3} = 32, 16, 8$, and depth range scaling factors $w_2, w_3 = \frac{1}{4}, \frac{1}{16}$ respectively. The loss weights for each stage $ \lambda_1,\lambda_2,\lambda_3 = 0.5,1,2 $. The network is trained for 160k iterations with a batch size of 2 by an Adam \cite{kingma2014adam} optimizer. The initial learning rate is 0.001 and is halved at the 100k, 120k and 140k steps. All experiments are performed using one Nvidia GTX 1080Ti card.

\vspace{-5mm}\paragraph{Point cloud generation}
Similar to previous works, we apply depth map filter and fusion approaches to merge all depth maps into a unified point cloud output. Both photometric and geometric consistencies are considered in our depth map filter and fusion step. For the photometric consistency, we follow \cite{yao2018mvsnet} and generate probability maps to filter out unreliable pixels. The summation of probabilities of depth hypothesis within range $[\mathbf{D}-2, \mathbf{D}+2]$ are calculated as the probability map of a given depth map output. Moreover, in our coarse-to-fine architecture, we consider all probability maps at different stages, and the filtering criterion is that a pixel in a reference view will be preserved if and only if all probability maps from all three stages are higher than the corresponding thresholds $p_{t,1}, p_{t,2}, p_{t,3}$. For geometric consistency, we preserve pixels whose depth estimation is consistent with the reprojected depth from at least $N_f$ views \cite{yao2018mvsnet}. Finally, the median depth map fusion is applied to refine all depth maps. The 3D point cloud is obtained by projecting all refined depth maps into the 3D space. 

\begin{table}[]
	\centering
	\resizebox{\textwidth}{!}{%
		\begin{tabular}{l|ccccccccc|ccc}
			\specialrule{.2em}{.1em}{.1em}
			& \multicolumn{9}{c|}{\textit{Tanks and Temples}} & \multicolumn{3}{c}{\textit{DTU (\textit{mm})}} \\
			& Mean & Family & Francis & Horse & Lighthouse & M60 & Panther & Playground & Train & Acc. & Comp. & Overall\\ \hline
			COLMAP \cite{schonberger2016pixelwise} & 42.14 & 50.41 & 22.25 & 25.63 & 56.43 & 44.83 & 46.97 & 48.53 & 42.04 & 0.400 & 0.664 & 0.532 \\ 
			MVSNet \cite{yao2018mvsnet} & 43.48 & 55.99 & 28.55 & 25.07 & 50.79 & 53.96 & 50.86 & 47.90 & 34.69 & 0.396 & 0.527 & 0.462 \\ 
			Point-MVSNet \cite{chen2019point} & 48.27 & 61.79 & 41.15 & 34.20 & 50.79 & 51.97 & 50.85 & 52.38 & 43.06 & 0.342 & 0.411 & 0.376 \\
			CVP-MVSNet \cite{yang2020cost} & 54.03 & 76.50 & 47.74 & 36.34 & 55.12 & 57.28 & 54.28 & 57.43 & 47.54 & \textbf{0.296} & 0.406 & 0.351 \\
			UCSNet \cite{cheng2020deep} & 54.83 & 76.09 & 53.16 & 43.03 & 54.00 & 55.60 & 51.49 & 57.38 & 47.89 & 0.338 & \textbf{0.349} & \textbf{0.344} \\
			CasMVSNet \cite{gu2020cascade} & 56.84 & 76.37 & 58.45 & 46.26 & 55.81 & 56.11 & 54.06 & 58.18 & 49.51 & 0.325 & 0.385 & 0.355 \\
			ACMM \cite{xu2019multi} & 57.27 & 69.24 & 51.45 & 46.97 & 63.20 & 55.07 & \textbf{57.64} & 60.08 & \textbf{54.48} & - & - & - \\ \hline
			Vis-MVSNet & \textbf{60.03} & \textbf{77.40} & \textbf{60.23} & \textbf{47.07} & \textbf{63.44} & \textbf{62.21} & 57.28 & \textbf{60.54} & 52.07 & 0.369 & 0.361 & 0.365 \\
			\specialrule{.2em}{.1em}{.1em}
		\end{tabular}
	}
	\caption{Quantitative result of the point cloud on the \textit{intermediate set} of \textit{Tanks and Temples} and the test set of \textit{DTU}. The proposed method achieves the best mean F-score among the listed works on \textit{Tanks and Temples} and comparable overall distance on \textit{DTU}. }
	\vspace{-4mm}
	\label{tab:tanks}
\end{table}

\subsection{Benchmarking on Tanks and Temples Dataset}\label{sec:tnt}

We first evaluate our method on the \textit{intermediate set} of \textit{Tanks and Temples} dataset \cite{knapitsch2017tanks}. As mentioned in Sec. \ref{sec:implementation}, we use the BlendedMVS training set \cite{yao2020blendedmvs} to train the network. BlendedMVS is a recent MVS dataset containing 113 indoor and outdoor scenes with 16904 MVS training samples in total. The dataset is split into 106 training scenes and 7 validation scenes. The trained model is directly applied to the \textit{Tanks and Temples} benchmarking without fine-tuning.

We use an input image size of $1920 \times 1080$ for reconstructions on the \textit{Tanks and Temples} dataset. The source image number is set to $N_v=7$ for network inference and we choose $N_f = 4$, $p_{t,1}, p_{t,2}, p_{t,3}=0.8, 0.7, 0.8$ for depth map filter and fusion. Quantitative results are shown in Tab.\ \ref{tab:tanks} and corresponding point cloud reconstructions are illustrated in Fig.\ \ref{fig:tanks}. Our Vis-MVSNet achieves a mean F-score of 60.03 and ranks $1^{st}$ among all the methods in the benchmark (until May 1, 2020), which outperforms all classical MVS methods \cite{schonberger2016pixelwise,xu2019multi} and recent learning-based approaches \cite{yao2018mvsnet,chen2019point,yang2020cost,cheng2020deep,gu2020cascade}.

\subsection{Benchmarking on DTU Dataset}\label{sec:dtu}

The proposed method is also benchmarked on the DTU evaluation set \cite{jensen2014large}. \textit{DTU} dataset contains 128 scans under fixed camera trajectories and 7 sets of lighting configuration. Every scan has 49 views with given camera parameters. As suggested by previous methods\cite{ji2017surfacenet,yao2018mvsnet}, DTU dataset is split into training set, validation set and evaluation set. Our model is trained on the DTU training set, which is mentioned in Sec. \ref{sec:implementation}

For the depth map estimation, we use an input image size of $1600\times1200$ and a fixed depth range of $[d_{min}, d_{max}] = [425mm, 905mm]$ for all input images. The source image number is set to $N_v=5$. We choose $N_f=2$ and $p_{t,1}, p_{t,2}, p_{t,3}=0.6, 0.6, 0.6$ for the depth map filter and fusion step. Quantitative results are shown in Tab.\ \ \ref{tab:tanks} and our method achieves a overall score of 0.365, which is comparable with other state-of-the-art methods. 

\begin{table}[]
	\begin{minipage}[b]{.55\linewidth}
		\strut\vspace*{-\baselineskip}\newline
		\centering
		\resizebox{\textwidth}{!}{%
			\begin{tabular}{l|c|ccc}
				\specialrule{.2em}{.1em}{.1em}
				Setting & Fusion Method & Loss & <1 (\%) & <3 (\%) \\ \hline
				base-var & Variance                  & 1.50     & 79.31          & 92.25          \\
				base-ave  & Average                  & 0.999     & 83.03          & 94.95          \\
				base-max  & Max Pooling                & 0.956     & 84.71          & 95.19          \\
				base-vis  & Proposed        & 0.908     & 85.35          & 95.48          \\ \hline
				proposed  & + Coarse-to-fine & 0.759     & 90.86          & 96.05     \\
				\specialrule{.2em}{.1em}{.1em}
		\end{tabular}}
		\vspace{-4mm}\caption{Quantitative result of the depth map on the validation set of \textit{BlendedMVS} with $ N_v=7 $. The settings with proposed fusion method achieve better results than others. }
		\label{tab:ablation}
	\end{minipage}
	\hspace{0.02\linewidth}
	\begin{minipage}[b]{.41\linewidth}
		\strut\vspace*{-\baselineskip}\newline
		\centering
		\includegraphics[width=\textwidth]{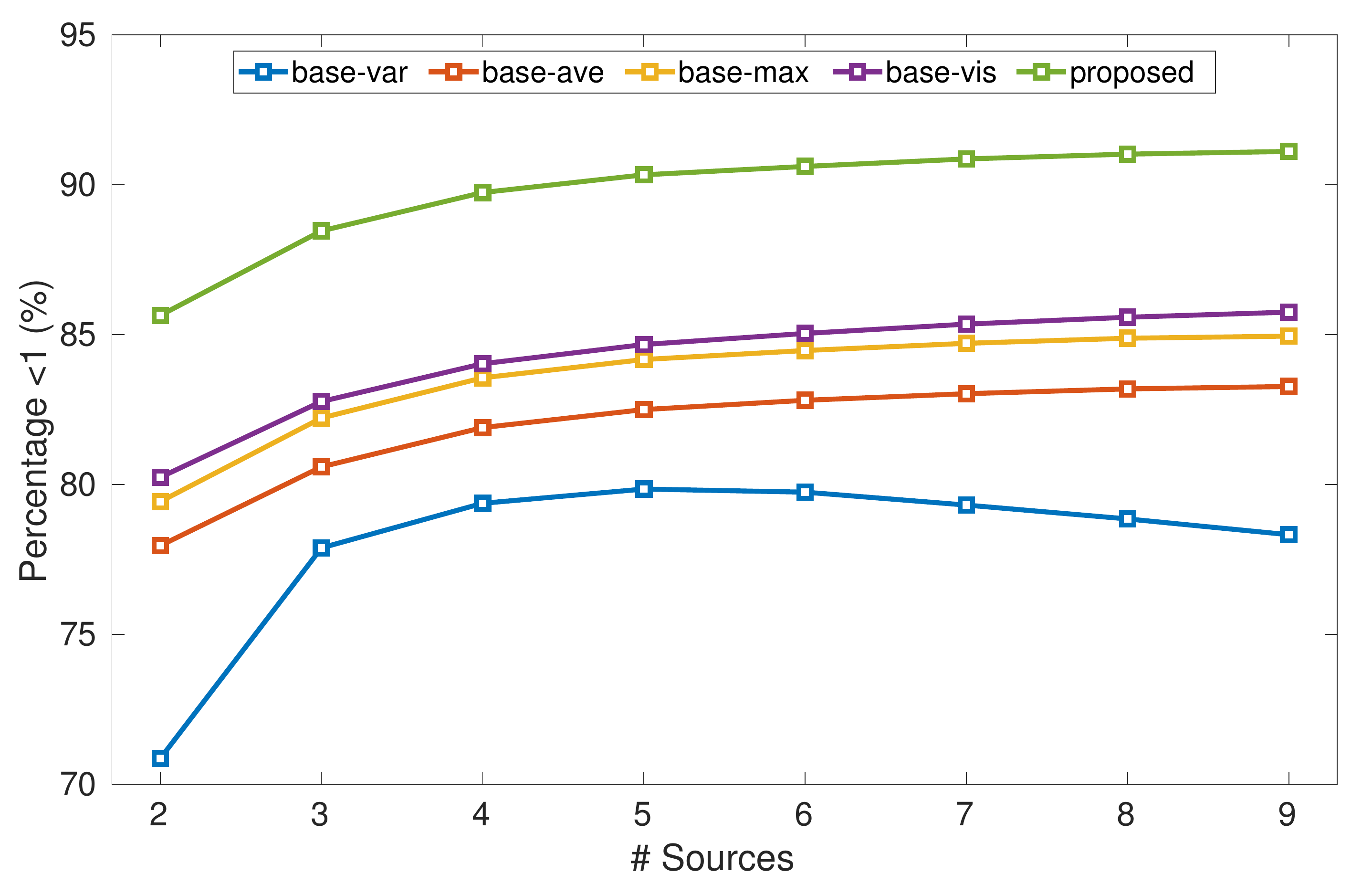}
		\vspace{-8mm}\captionof{figure}{Percentage of <1 of the depth maps on \textit{BlendedMVS} w.r.t. $ N_v $. }\vspace{-3mm}
		\label{fig:ablation}
	\end{minipage}
\end{table}

\subsection{Ablation Study}\label{sec:ablation}
In this section, we discuss other alternative volume fusion methods with implicit or explicit visibility awareness. To keep the simplicity of the network and clear demonstrate the effectiveness of the proposed component, we remove the coarse-to-fine architecture and directly use a MVSNet-like network as our baseline. The ablation study is performed on the BlendedMVS validation set and three types of evaluation metrics are considered: 1) the average L1 loss between the inferred depth map and the ground truth depth map; 2) the percentage of pixels with L1 error smaller than 1 depth-wise pixel ($< 1$ percentage); and 3) the <3 percentage. Quantitative results are shown in Tab.\ \ref{tab:ablation} and Fig. \ref{fig:ablation}

\vspace{-4mm}\paragraph{Baseline}
In this setting (\textit{base-var}), we directly use the variance metric to fuse the feature volumes into one cost volume. The \textit{base-var} setting is widely adopted by MVSNet and its following works \cite{yao2018mvsnet,chen2019point,yang2020cost,cheng2020deep,gu2020cascade}. However, the variance operation is under the assumption that all pixels in the reference should be visible from all views. As a result, the increasing input image number would lead to even worse evaluation metrics (see Fig. \ref{fig:ablation})

\vspace{-4mm}\paragraph{Averaging}
In this setting (\textit{base-ave}), pair-wise cost volumes are fused to one multi-view volume by direct element-wise averaging. To fairly compare this setting with the proposed setting, we also apply the two step regularization as in the proposed framework. As is shown in Fig. \ref{fig:ablation}, the <1 percentage accuracy of the \textit{base-ave} is consistently increasing with the input image number. We believe the visibility information is implicitly encoded in the latent space and is dealt with by the two-step regularization. However, such implicit visibility awareness is apparently inferior to the proposed visibility fusion approach (see \textit{base-vis} in Tab.\ \ref{tab:ablation} and Fig. \ref{fig:ablation}).

\vspace{-4mm}\paragraph{Max Pooling}
In this setting (\textit{base-max}), the fused volume is obtained by finding the element-wise maximum of all the pair-wise volumes. This setting follows the fusion strategy of only considering the best matching pair among all reference-source image pairs. Similarly, all pair-wise losses are not counted toward the final loss. As is shown in Tab.\ \ref{tab:ablation} and Fig. \ref{fig:ablation}, \textit{base-max} outperforms \textit{base-ave} but is still inferior to the proposed \textit{base-vis}.

\vspace{-4mm}\paragraph{Weighted Averaging}
This setting (\textit{base-vis}) is the proposed Vis-MVSNet without the coarse-to-fine architecture. Compared with \textit{base-ave} and \textit{base-max}, this setting utilizes the intermediate uncertainty as the weighting guidance for the pair-wise volume fusion. As the result, the significance of invisible pixels will be explicitly reduced in the volume fusion step. 

The quantitative comparison is shown in Tab.\ \ref{tab:ablation} and Fig.\ \ref{fig:ablation}. A significant improvement can be observed after introducing the two step regularization to the baseline (\textit{base-ave} and \textit{base-max} v.s.\ \textit{base-var}). In addition, the proposed fusion further improves the result (\textit{base-vis} v.s.\ \textit{base-ave} and \textit{base-max}). Finally, the full model with coarse-to-fine architecture outperforms others by a significant margin (\textit{proposed} v.s.\ others).

\section{Conclusion}
We have presented a visibility-aware depth inference framework for multi-view stereo reconstruction. We have proposed the two-step cost volume regularization, the joint inference of the pair-wise depth and the uncertainty, and the weighted average fusion of pair-wise volumes according to the uncertainty maps. The proposed method has been extensively evaluated on several datasets,  demonstrating the effectiveness of the proposed visibility-aware depth inference framework.

\section{Acknowledgments}
This work is supported by Hong Kong RGC GRF 16206819 \& 16203518 and T22-603/15N.

\bibliography{egbib}

\begin{thebibliography}{33}
\providecommand{\natexlab}[1]{#1}
\providecommand{\url}[1]{\texttt{#1}}
\expandafter\ifx\csname urlstyle\endcsname\relax
  \providecommand{\doi}[1]{doi: #1}\else
  \providecommand{\doi}{doi: \begingroup \urlstyle{rm}\Url}\fi

\bibitem[Campbell et~al.(2008)Campbell, Vogiatzis, Hern{\'a}ndez, and
  Cipolla]{campbell2008using}
Neill~DF Campbell, George Vogiatzis, Carlos Hern{\'a}ndez, and Roberto Cipolla.
\newblock Using multiple hypotheses to improve depth-maps for multi-view
  stereo.
\newblock In \emph{European Conference on Computer Vision (ECCV)}, 2008.

\bibitem[Chen et~al.(2019)Chen, Han, Xu, and Su]{chen2019point}
Rui Chen, Songfang Han, Jing Xu, and Hao Su.
\newblock Point-based multi-view stereo network.
\newblock In \emph{International Conference on Computer Vision (ICCV)}, 2019.

\bibitem[Cheng et~al.(2020)Cheng, Xu, Zhu, Li, Li, Ramamoorthi, and
  Su]{cheng2020deep}
Shuo Cheng, Zexiang Xu, Shilin Zhu, Zhuwen Li, Li~Erran Li, Ravi Ramamoorthi,
  and Hao Su.
\newblock Deep stereo using adaptive thin volume representation with
  uncertainty awareness.
\newblock In \emph{Computer Vision and Pattern Recognition (CVPR)}, 2020.

\bibitem[Furukawa and Ponce(2009)]{furukawa2009accurate}
Yasutaka Furukawa and Jean Ponce.
\newblock Accurate, dense, and robust multiview stereopsis.
\newblock \emph{IEEE transactions on pattern analysis and machine
  intelligence}, 32\penalty0 (8):\penalty0 1362--1376, 2009.

\bibitem[Galliani et~al.(2015)Galliani, Lasinger, and
  Schindler]{galliani2015massively}
Silvano Galliani, Katrin Lasinger, and Konrad Schindler.
\newblock Massively parallel multiview stereopsis by surface normal diffusion.
\newblock In \emph{International Conference on Computer Vision (ICCV)}, 2015.

\bibitem[Gu et~al.(2020)Gu, Fan, Zhu, Dai, Tan, and Tan]{gu2020cascade}
Xiaodong Gu, Zhiwen Fan, Siyu Zhu, Zuozhuo Dai, Feitong Tan, and Ping Tan.
\newblock Cascade cost volume for high-resolution multi-view stereo and stereo
  matching.
\newblock In \emph{Computer Vision and Pattern Recognition (CVPR)}, 2020.

\bibitem[Guo et~al.(2019)Guo, Yang, Yang, Wang, and Li]{guo2019group}
Xiaoyang Guo, Kai Yang, Wukui Yang, Xiaogang Wang, and Hongsheng Li.
\newblock Group-wise correlation stereo network.
\newblock In \emph{Computer Vision and Pattern Recognition (CVPR)}, 2019.

\bibitem[Hartmann et~al.(2017)Hartmann, Galliani, Havlena, Van~Gool, and
  Schindler]{hartmann2017learned}
Wilfried Hartmann, Silvano Galliani, Michal Havlena, Luc Van~Gool, and Konrad
  Schindler.
\newblock Learned multi-patch similarity.
\newblock In \emph{International Conference on Computer Vision (ICCV)}, 2017.

\bibitem[Hu and Mordohai(2012)]{hu2012quantitative}
Xiaoyan Hu and Philippos Mordohai.
\newblock A quantitative evaluation of confidence measures for stereo vision.
\newblock \emph{IEEE transactions on pattern analysis and machine
  intelligence}, 34\penalty0 (11):\penalty0 2121--2133, 2012.

\bibitem[Huang et~al.(2018)Huang, Matzen, Kopf, Ahuja, and
  Huang]{huang2018deepmvs}
Po-Han Huang, Kevin Matzen, Johannes Kopf, Narendra Ahuja, and Jia-Bin Huang.
\newblock Deepmvs: Learning multi-view stereopsis.
\newblock In \emph{Computer Vision and Pattern Recognition (CVPR)}, 2018.

\bibitem[Jensen et~al.(2014)Jensen, Dahl, Vogiatzis, Tola, and
  Aan{\ae}s]{jensen2014large}
Rasmus Jensen, Anders Dahl, George Vogiatzis, Engil Tola, and Henrik Aan{\ae}s.
\newblock Large scale multi-view stereopsis evaluation.
\newblock In \emph{Computer Vision and Pattern Recognition (CVPR)}, 2014.

\bibitem[Ji et~al.(2017)Ji, Gall, Zheng, Liu, and Fang]{ji2017surfacenet}
Mengqi Ji, Juergen Gall, Haitian Zheng, Yebin Liu, and Lu~Fang.
\newblock Surfacenet: An end-to-end 3d neural network for multiview stereopsis.
\newblock In \emph{International Conference on Computer Vision (ICCV)}, 2017.

\bibitem[Kar et~al.(2017)Kar, H{\"a}ne, and Malik]{kar2017learning}
Abhishek Kar, Christian H{\"a}ne, and Jitendra Malik.
\newblock Learning a multi-view stereo machine.
\newblock In \emph{Advances in neural information processing systems}, 2017.

\bibitem[Kendall and Gal(2017)]{kendall2017uncertainties}
Alex Kendall and Yarin Gal.
\newblock What uncertainties do we need in bayesian deep learning for computer
  vision?
\newblock In \emph{Advances in neural information processing systems}, 2017.

\bibitem[Kendall et~al.(2017)Kendall, Martirosyan, Dasgupta, Henry, Kennedy,
  Bachrach, and Bry]{kendall2017end}
Alex Kendall, Hayk Martirosyan, Saumitro Dasgupta, Peter Henry, Ryan Kennedy,
  Abraham Bachrach, and Adam Bry.
\newblock End-to-end learning of geometry and context for deep stereo
  regression.
\newblock In \emph{International Conference on Computer Vision (ICCV)}, 2017.

\bibitem[Kim et~al.(2018)Kim, Min, Kim, and Sohn]{kim2018unified}
Sunok Kim, Dongbo Min, Seungryong Kim, and Kwanghoon Sohn.
\newblock Unified confidence estimation networks for robust stereo matching.
\newblock \emph{IEEE Transactions on Image Processing}, 28\penalty0
  (3):\penalty0 1299--1313, 2018.

\bibitem[Kim et~al.(2019)Kim, Kim, Min, and Sohn]{kim2019laf}
Sunok Kim, Seungryong Kim, Dongbo Min, and Kwanghoon Sohn.
\newblock Laf-net: Locally adaptive fusion networks for stereo confidence
  estimation.
\newblock In \emph{Computer Vision and Pattern Recognition (CVPR)}, 2019.

\bibitem[Kingma and Ba(2014)]{kingma2014adam}
Diederik~P Kingma and Jimmy Ba.
\newblock Adam: A method for stochastic optimization.
\newblock \emph{arXiv preprint arXiv:1412.6980}, 2014.

\bibitem[Knapitsch et~al.(2017)Knapitsch, Park, Zhou, and
  Koltun]{knapitsch2017tanks}
Arno Knapitsch, Jaesik Park, Qian-Yi Zhou, and Vladlen Koltun.
\newblock Tanks and temples: Benchmarking large-scale scene reconstruction.
\newblock \emph{ACM Transactions on Graphics (ToG)}, 36\penalty0 (4):\penalty0
  78, 2017.

\bibitem[Paschalidou et~al.(2018)Paschalidou, Ulusoy, Schmitt, Van~Gool, and
  Geiger]{paschalidou2018raynet}
Despoina Paschalidou, Osman Ulusoy, Carolin Schmitt, Luc Van~Gool, and Andreas
  Geiger.
\newblock Raynet: Learning volumetric 3d reconstruction with ray potentials.
\newblock In \emph{Computer Vision and Pattern Recognition (CVPR)}, 2018.

\bibitem[Poggi and Mattoccia(2016)]{poggi2016learning}
Matteo Poggi and Stefano Mattoccia.
\newblock Learning from scratch a confidence measure.
\newblock In \emph{British Machine Vision Conference (BMVC)}, 2016.

\bibitem[Ronneberger et~al.(2015)Ronneberger, Fischer, and
  Brox]{ronneberger2015u}
Olaf Ronneberger, Philipp Fischer, and Thomas Brox.
\newblock U-net: Convolutional networks for biomedical image segmentation.
\newblock In \emph{Medical image computing and computer-assisted intervention},
  2015.

\bibitem[Sch{\"o}nberger et~al.(2016)Sch{\"o}nberger, Zheng, Frahm, and
  Pollefeys]{schonberger2016pixelwise}
Johannes~L Sch{\"o}nberger, Enliang Zheng, Jan-Michael Frahm, and Marc
  Pollefeys.
\newblock Pixelwise view selection for unstructured multi-view stereo.
\newblock In \emph{European Conference on Computer Vision (ECCV)}, 2016.

\bibitem[Tola et~al.(2012)Tola, Strecha, and Fua]{tola2012efficient}
Engin Tola, Christoph Strecha, and Pascal Fua.
\newblock Efficient large-scale multi-view stereo for ultra high-resolution
  image sets.
\newblock \emph{Machine Vision and Applications}, 23\penalty0 (5):\penalty0
  903--920, 2012.

\bibitem[Tosi et~al.(2018)Tosi, Poggi, Benincasa, and
  Mattoccia]{tosi2018beyond}
Fabio Tosi, Matteo Poggi, Antonio Benincasa, and Stefano Mattoccia.
\newblock Beyond local reasoning for stereo confidence estimation with deep
  learning.
\newblock In \emph{European Conference on Computer Vision (ECCV)}, 2018.

\bibitem[Xu and Tao(2019)]{xu2019multi}
Qingshan Xu and Wenbing Tao.
\newblock Multi-scale geometric consistency guided multi-view stereo.
\newblock In \emph{Computer Vision and Pattern Recognition (CVPR)}, 2019.

\bibitem[Xue et~al.(2019)Xue, Chen, Wan, Huang, Yu, Li, and Bao]{xue2019mvscrf}
Youze Xue, Jiansheng Chen, Weitao Wan, Yiqing Huang, Cheng Yu, Tianpeng Li, and
  Jiayu Bao.
\newblock Mvscrf: Learning multi-view stereo with conditional random fields.
\newblock In \emph{International Conference on Computer Vision (ICCV)}, 2019.

\bibitem[Yang et~al.(2020)Yang, Mao, Alvarez, and Liu]{yang2020cost}
Jiayu Yang, Wei Mao, Jose~M Alvarez, and Miaomiao Liu.
\newblock Cost volume pyramid based depth inference for multi-view stereo.
\newblock In \emph{Computer Vision and Pattern Recognition (CVPR)}, 2020.

\bibitem[Yao et~al.(2018)Yao, Luo, Li, Fang, and Quan]{yao2018mvsnet}
Yao Yao, Zixin Luo, Shiwei Li, Tian Fang, and Long Quan.
\newblock Mvsnet: Depth inference for unstructured multi-view stereo.
\newblock In \emph{European Conference on Computer Vision (ECCV)}, 2018.

\bibitem[Yao et~al.(2019)Yao, Luo, Li, Shen, Fang, and Quan]{yao2019recurrent}
Yao Yao, Zixin Luo, Shiwei Li, Tianwei Shen, Tian Fang, and Long Quan.
\newblock Recurrent mvsnet for high-resolution multi-view stereo depth
  inference.
\newblock In \emph{Computer Vision and Pattern Recognition (CVPR)}, 2019.

\bibitem[Yao et~al.(2020)Yao, Luo, Li, Zhang, Ren, Zhou, Fang, and
  Quan]{yao2020blendedmvs}
Yao Yao, Zixin Luo, Shiwei Li, Jingyang Zhang, Yufan Ren, Lei Zhou, Tian Fang,
  and Long Quan.
\newblock Blendedmvs: A large-scale dataset for generalized multi-view stereo
  networks.
\newblock In \emph{Computer Vision and Pattern Recognition (CVPR)}, 2020.

\bibitem[Zhang et~al.(2020)Zhang, Yao, Luo, Li, Shen, Fang, and
  Quan]{zhang2020learning}
Jingyang Zhang, Yao Yao, Zixin Luo, Shiwei Li, Tianwei Shen, Tian Fang, and
  Long Quan.
\newblock Learning stereo matchability in disparity regression networks.
\newblock \emph{arXiv preprint arXiv:2008.04800}, 2020.

\bibitem[Zheng et~al.(2014)Zheng, Dunn, Jojic, and Frahm]{zheng2014patchmatch}
Enliang Zheng, Enrique Dunn, Vladimir Jojic, and Jan-Michael Frahm.
\newblock Patchmatch based joint view selection and depthmap estimation.
\newblock In \emph{Computer Vision and Pattern Recognition (CVPR)}, 2014.

\end{thebibliography}
\end{document}